\documentclass[conference]{IEEEtran}
\IEEEoverridecommandlockouts

\usepackage{cite}
\usepackage{amsmath,amssymb,amsfonts}
\usepackage{algorithmic}
\usepackage{graphicx}
\usepackage{textcomp}
\usepackage{xspace}
\usepackage{xcolor}
\usepackage{booktabs}
\usepackage{multirow}
\usepackage{subcaption} 
\usepackage{pifont}
\usepackage{float} 
\usepackage{makecell}
\usepackage{url}
\usepackage[most]{tcolorbox}

\usepackage{newfloat}

\def\BibTeX{{\rm B\kern-.05em{\sc i\kern-.025em b}\kern-.08em
    T\kern-.1667em\lower.7ex\hbox{E}\kern-.125emX}}

\newcommand{\ourmethod}{HOPE\xspace}
\begin{document}

\title{HOPE: Heterophily-Aware Open-Set Node Classification with Pseudo-Extrapolation}

\author{
\IEEEauthorblockN{
Yumeng Dai\textsuperscript{1},
Yue Tan\textsuperscript{2},
Yixin Liu\textsuperscript{2},
Chenxu Wang\textsuperscript{1,*},
Pinghui Wang\textsuperscript{1},
Tao Qin\textsuperscript{1}
}

\IEEEauthorblockA{
\textsuperscript{1}Xi'an Jiaotong University, Xi'an, China\\
\textsuperscript{2}Griffith University, Brisbane, Australia\\
ymdai@stu.xjtu.edu.cn,
yue.tan@griffith.edu.au,
yixin.liu@griffith.edu.au,\\
cxwang@mail.xjtu.edu.cn,
phwang@mail.xjtu.edu.cn,
qin.tao@mail.xjtu.edu.cn\\
\textsuperscript{*}Corresponding author: Chenxu Wang
}
}


\maketitle

\begin{abstract}
Standard open-set node classification methods heavily rely on the homophily assumption, where connected nodes share identical labels. 
However, real-world graphs are often heterophilic, revealing the sub-optimal performance of current methods and posing new challenges to open-set node classification. 
On the one hand, the cross-class connectivity nature of heterophilic graphs causes node representations from different known or unknown classes to be intertwined after aggregation, undermining the discriminative capacity of learned representations. 
On the other hand, the structural mixture invalidates traditional threshold-based open-set classification methods and breaks conventional cross-class feature interpolation paradigms, leading to unreliable unknown-class rejection. 
To address these critical challenges, we propose a novel framework named \ourmethod \underline{H}eterophily-aware \underline{O}pen-set node classification method with \underline{P}seudo-\underline{E}xtrapolation, abbreviated as HOPE. To adapt open-set graph neural networks (GNNs) to heterophilic scenarios, \ourmethod utilizes a structure-augmented feature initialization layer to capture multi-hop structural patterns for feature augmentation. Meanwhile, we design a trustworthy neighborhood aggregation mechanism adaptable to standard GNNs to dynamically filter out noisy cross-class neighbors. 
To enhance the unknown-class rejection capability of \ourmethod, we introduce a heterophily-guided pseudo-extrapolation strategy. It dynamically maintains the representations of known-class centers and extrapolates from these centers along cross-class neighborhood displacement directions, thereby synthesizing pseudo-unknown proxies near structurally ambiguous regions.
Finally, we optimize the network via a joint classification framework with logit margin regularization, routing synthetic proxies into a dedicated rejection slot without imposing geometric margin constraints in the representation space. 
Extensive experiments on multiple datasets demonstrate that \ourmethod consistently outperforms state-of-the-art models, validating its effectiveness, robustness, and efficiency.
\end{abstract}


\section{Introduction}

Graph Neural Networks (GNNs) have achieved remarkable success in various graph-structured tasks, particularly in semi-supervised node classification \cite{kipf2016semi}. Standard node classification methods operate under a closed-set assumption, where the training and testing environments share an identical label space. 
However, real-world graph database applications often encounter open-set scenarios \cite{wu2020openwgl, gao2024graph,liu2026few}, where test graphs contain nodes from previously unseen or unknown categories. 
When conventional closed-set methods face these unknown nodes, they often mistakenly force them into one of the known classes with high confidence, neglecting the presence of novel categories beyond the predefined label space. To address this limitation, \textbf{open-set node classification} has emerged as a crucial research problem, aiming to classify nodes from known categories while identifying nodes from previously unseen categories~\cite{huang2022end, wu2020openwgl}. 
By jointly modeling known-class discrimination and unknown-class rejection, open-set node classification methods can support both accurate known-class identification and robust unknown-class detection.

\begin{figure}[t]
    \centering
    \subfloat[Unknown classes in homophilic graphs]{\includegraphics[width=0.48\linewidth]{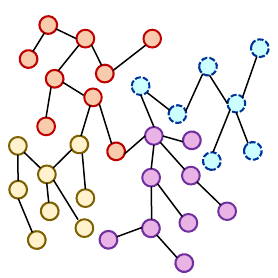}%
    \label{figs:homo}}
    \hfill
    \subfloat[Unknown categories in heterophilic graphs]{\includegraphics[width=0.48\linewidth]{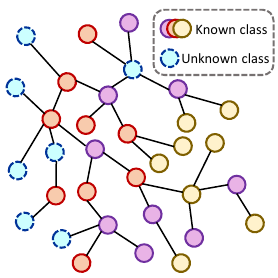}%
    \label{figs:heter}}
    \caption{The distinct topological behaviors of unknown class nodes in homophilic versus heterophilic graphs.}
    \label{figs:compare}
\end{figure}
Despite the progress of open-set node classification in recent years, existing methods often assume that graphs are homophilic, meaning that connected nodes tend to share the same class label~\cite{zheng2026graph}. This assumption may not always hold in practical scenarios, since heterophilic graphs, which break the above homophily assumption, are ubiquitous in practical domains, such as e-commerce networks, social networks, and molecular structures \cite{zheng2026graph, lin2025heterophily, wang2022powerful, zhu2020beyond, tan2026influence, chen2025multi}. 
Due to their distinct neighborhood-label correlations, the structural distribution of unknown classes exhibits fundamental differences between homophilic and heterophilic settings. As shown in Fig.~\ref{figs:homo}, unseen categories in homophilic graphs are characterized by dense localized clustering with limited cross-boundary expansion. Conversely, as depicted in Fig.~\ref{figs:heter}, unknown nodes in heterophilic graphs manifest as highly intertwined topological neighbors, generating extensive and complex mixtures that deeply blur the classification boundaries of adjacent known class manifolds. 
As a result, existing open-set node classification methods often struggle on heterophilic graphs, as their homophily-oriented mechanisms, such as label propagation~\cite{iscen2019label, wang2020unifying} and consistency regularization~\cite{bo2022regularizing, li2021comatch}, may over-smooth naturally dissimilar neighbors and misinterpret normal inter-class edges as label noise. Moreover, the boundary-based proxy generation strategies (e.g., Mixup) used in some open-set classification approaches~\cite{pxy_zhang2023g2pxy, zhang2024egonc} may scatter pseudo-unknown nodes across the feature space in heterophilic graphs, thereby blurring the distinction between known and unknown classes. 
These limitations motivate a fundamental question:

\begin{tcolorbox}[
    colback=gray!8,
    colframe=gray!40,
    boxrule=0.5pt,
    arc=1.5mm,
    left=1mm,
    right=1mm,
    top=0.5mm,
    bottom=0.5mm,
    boxsep=1mm
]
\centering
\textbf{\textit{Can we develop an open-set graph learning approach that can identify unknown nodes in heterophilic graphs?}}
\end{tcolorbox}

Answering the above question is non-trivial, as the structural discrepancies between homophilic and heterophilic graphs introduce two key challenges for open-set node classification. \textit{\textbf{Challenge 1}: Representation boundary confusion under heterophilic aggregation.} 
In a heterophilic graph, a known-class node is surrounded by neighbors from different known classes and potential unknown classes \cite{liu2025integrating,shen2026raising}. In this case, standard neighborhood aggregation, which acts as a low-pass filter, may smooth out essential high-frequency distinctive features. When open-set nodes are mixed into these neighborhoods, uniform aggregation can cause catastrophic representation overlap, making known-class discrimination and unknown-class rejection mutually entangled. As a result, the model struggles to separate different known classes in the representation space, since the neighbor label distributions are highly noisy and complex, leading to ambiguous decision boundaries among the known classes.

While the above challenge focuses on maintaining separable known-class representations, open-set learning further requires reliable rejection of nodes beyond the known label space, leading to \textit{\textbf{Challenge 2}: Unreliable unknown-class rejection under structural entanglement.}
In heterophilic graphs, unknown nodes are structurally intertwined with the clusters from known classes through cross-class connections, rendering existing rejection criteria for open-set node classification ineffective. 
Specifically, threshold-based rejection methods~\cite{wu2020openwgl, zhang2024rog_pl, zhang2024conc} may misinterpret heterophilic connections as unknown evidence, leading to both false positives and false negatives. 
Meanwhile, interpolation-based proxy generation methods~\cite{pxy_zhang2023g2pxy, zhang2024egonc} may also become unreliable, as simple feature interpolation between known classes cannot faithfully capture the complex cross-class distributions and topological boundaries induced by heterophily. Consequently, the generated proxies may be scattered across the feature space, further misleading the model in separating unknown nodes from known ones. 
In this case, a dedicated open-set node classification framework is needed to detect unknown nodes under such structurally entangled heterophilic settings.

To address these challenges, we propose a novel \textbf{H}eterophily-aware \textbf{O}pen-set node classification method with \textbf{P}seudo-\textbf{E}xtrapolation (\ourmethod for short). \ourmethod consists of four components designed to support known-class node classification and unknown-class node rejection simultaneously. Specifically, to address \textbf{\textit{Challenge 1}}, we develop a structure-augmented feature initialization layer that computes multi-hop geometric descriptors to enrich raw node attributes, ensuring the model recognizes latent open-set boundaries before message passing. Meanwhile, we implement a trustworthy neighborhood aggregation scheme that uses an edge discriminator to evaluate homophily compatibility scores and dynamically prunes untrustworthy cross-class neighbors. Instead of relying on a specific GNN backbone, the proposed aggregation scheme can be applied to various standard GNN models in a plug-and-play manner, enhancing their ability to preserve discriminative known-class representations under heterophily. To handle \textbf{\textit{Challenge 2}}, 
we introduce a pseudo-unknown proxy generation strategy based on structural extrapolation. 
This strategy extrapolates from known-class anchors along heterophilic neighborhood directions to synthesize representative pseudo-unknown boundary instances, 
providing informative supervision for learning reliable known-unknown rejection boundaries. 
Moreover, we optimize the network via a unified classification framework integrated with a logit margin regularization loss. By routing the synthetic proxies to a dedicated $(K+1)$-th classification slot, \ourmethod establishes a robust classification-driven shield around known categories without relying on explicit geometric margin constraints in the representation space. 
The primary contributions of this paper are summarized as follows:
\begin{itemize}
    \item \textbf{Problem:} We formalize the problem of open-set node classification under heterophily and investigate the topological distribution differences of unknown classes between homophilic and heterophilic graphs.
    \item \textbf{Methodology:} We propose \ourmethod, which incorporates a structure-augmented initialization layer, a trustworthy neighbor filtering mechanism adaptable to standard GNNs, a heterophily-guided pseudo-extrapolation strategy, and a unified $(K+1)$-way classification framework with logit regularization.
    \item \textbf{Experiments:} Extensive evaluations on multiple heterophilic graph benchmarks demonstrate that \ourmethod significantly outperforms state-of-the-art closed-set and open-set models, showcasing superior generalizability and robustness.
\end{itemize}

\section{Related Work}
\label{sec:related_work}

\subsection{Graph Neural Networks under Heterophily}
Most foundational Graph Neural Networks (GNNs), such as GCN \cite{kipf2016semi} and GAT \cite{velivckovic2017graph}, assume graph homophily, where connected nodes share similar labels or features \cite{zheng2026graph}. However, these models fail on heterophilic graphs where links connect nodes from different classes, a pattern common in e-commerce and social networks \cite{zheng2026graph, lin2025heterophily, wang2022powerful, zhu2020beyond,zhao2026fedcigar, li2026relationalAD}. 

To capture heterophilic graph structures, various message-passing and spectral mechanisms have been designed. For instance, H2GCN \cite{zhu2020beyond} separates ego-embeddings from neighborhood representations, Geom-GCN \cite{pei2020geom} maps topologies into geometric spaces, and JK-Net \cite{xu2018representation} aggregates layers dynamically. Additionally, GPR-GNN \cite{chien2020adaptive} utilizes generalized PageRank for adaptive filtering, GCNII \cite{chen2020simple} incorporates initial residuals, EG-GCN \cite{liu2025integrating} employs edge discriminators, and DiRW \cite{su2025dirw} uses path-aware random walks. Yet, these methods assume a fixed and fully known label space during training, causing them to fail when encountering unknown classes at test time.

\subsection{Open-Set Node Classification on Graphs}
Open-set node classification addresses this by identifying unknown semantic classes during inference \cite{wu2020openwgl, xu2024lego, tan2023taming}, primarily through threshold-based calibration or generative proxy modeling. Threshold-based frameworks like OpenWGL \cite{wu2020openwgl} utilize rejection metrics on softmax confidence or uncertainty scores \cite{zhang2024rog_pl, zhang2024conc}, but deep networks often output overconfident scores for unknown samples \cite{gawlikowski2023survey}. Generative proxy modeling methods, such as G2Pxy \cite{pxy_zhang2023g2pxy} and EGonc \cite{zhang2024egonc}, instead introduce virtual open-set nodes via hidden-layer manifold mixup or energy-based density optimization to simulate external distributions.

Nevertheless, existing open-set methods strongly depend on structural homophily, assuming unknown categories always appear as localized, tight groups. Under severe heterophily, known and unknown nodes connect tightly, causing standard propagation to blur semantic spaces and create severe overlap at classification boundaries. To address this, \ourmethod\ introduces a classification-driven structural extrapolation mechanism optimized within local batches. By mapping synthetic boundary proxies to a dedicated $(K+1)$-th classification slot, this framework maintains high closed-set accuracy while ensuring adaptive open-set node rejection on heterophilic graphs.

\section{Preliminaries}
\label{sec:preliminaries}

In this section, we present the formal definitions of graph concepts, heterophily, and the formulation of the open-set node classification task on heterophilic graphs. 

\subsection{Graph Definitions and Heterophily}
Let $\mathcal{G} = (\mathcal{V}, \mathcal{E}, \mathbf{X})$ denote an attributed graph, where $\mathcal{V} = \{v_1, v_2, \dots, v_N\}$ represents the set of $N$ nodes, and $\mathcal{E} \subseteq \mathcal{V} \times \mathcal{V}$ represents the set of edges. 
The topological structure of $\mathcal{G}$ can be uniquely represented by an adjacency matrix $\mathbf{A} \in \{0, 1\}^{N \times N}$, where $\mathbf{A}_{ij} = 1$ if there exists an edge $(v_i, v_j) \in \mathcal{E}$, and $\mathbf{A}_{ij} = 0$ otherwise. 
Each node $v_i \in \mathcal{V}$ is associated with a $D$-dimensional feature vector $\mathbf{x}_i \in \mathbb{R}^D$, and the collective features of all nodes form the node attribute matrix $\mathbf{X} = [\mathbf{x}_1, \mathbf{x}_2, \dots, \mathbf{x}_N]^\top \in \mathbb{R}^{N \times D}$. 
The neighborhood of a node $v_i$ is defined as $\mathcal{N}(v_i) = \{v_j \in \mathcal{V} \mid (v_i, v_j) \in \mathcal{E}\}$.

The connection patterns between nodes in a graph can be quantified by homophily and heterophily. Formally, given a fully labeled graph where each node $v_i$ has a class label $y_i$, the edge homophily ratio $h$ is defined as the proportion of edges that connect nodes sharing the same label:
\begin{equation}
    h = \frac{\sum_{(v_i, v_j) \in \mathcal{E}} \mathbb{I}(y_i = y_j)}{|\mathcal{E}|},
\end{equation}
where $\mathbb{I}(\cdot)$ is the indicator function. 
A graph is conventionally categorized as a \textit{homophilic graph} when $h$ is close to 1, implying that ``like attracts like''. Conversely, a graph is designated as a \textit{heterophilic graph} when $h$ is close to 0, which signifies that edges predominantly link nodes belonging to distinct categories (i.e., $\mathcal{N}(v_i)$ contains substantial semantic diversity).

\subsection{Open-Set Node Detection Formulation}
Unlike classical closed-set semi-supervised node classification, where the training and testing phases share an identical label space, Open-Set Recognition (OSR) accommodates the presence of unknown semantic classes during inference. 
Formally, let $\mathcal{Y}_L = \{c_1, c_2, \dots, c_K\}$ be the set of $K$ known classes available during the training stage. 
In the open-set deployment phase, the test nodes may originate from an expanded label space $\mathcal{Y} = \mathcal{Y}_L \cup \mathcal{Y}_U$, where $\mathcal{Y}_U = \{c_{K+1}, c_{K+2}, \dots\}$ denotes the set of unseen or unknown classes that never appear in the training dataset, satisfying $\mathcal{Y}_L \cap \mathcal{Y}_U = \emptyset$.

During training, we are given the graph $\mathcal{G}$ along with a set of labeled training nodes $\mathcal{V}_{train} \subset \mathcal{V}$, where each node $v_i \in \mathcal{V}_{train}$ is assigned a known label $y_i \in \mathcal{Y}_L$. 
The remaining nodes are partitioned into a validation set $\mathcal{V}_{val}$ and a test set $\mathcal{V}_{test}$. 
Critically, $\mathcal{V}_{test}$ comprises both known-class nodes (whose labels belong to $\mathcal{Y}_L$) and unknown-class nodes (whose true labels belong to $\mathcal{Y}_U$). 
The objective of open-set node detection is to learn a mapping function $\mathcal{F}: \mathcal{V} \rightarrow \mathcal{Y}_L \cup \{c_{unk}\}$, capable of precisely classifying nodes from known classes into their respective categories while simultaneously rejecting nodes from any unseen classes by categorizing them into a single unified unknown slot $c_{unk}$ (typically mapped as the $(K+1)$-th class).

\subsection{Open-Set GNNs under Heterophily}
An Open-Set Graph Neural Network (GNN) model typically consists of a structural encoder followed by an open-set classifier. 
Existing traditional open-set learning paradigms frequently assume that data samples are independent and identically distributed (i.i.d.). 
Open-set GNNs break this assumption by leveraging spatial dependencies, propagating ego-features across the topological structure via message-passing mechanisms to generate robust node representations $\mathbf{Z} = \text{Encoder}(\mathbf{X}, \mathbf{A})$. 

However, in heterophilic environments, standard open-set GNNs that rely on uniform low-pass aggregation (such as GCN) inherently aggregate conflicting representations from dissimilar neighbors. 
This drawback distorts the open-set decision boundaries by mixing known and unknown semantic spaces in the neighborhood. 
To resolve this challenge, \ourmethod\ addresses the coupled heterophily and open-set configuration by learning heterophily-aware structural mappings that separate multi-hop structural patterns. 
Instead of relying on representation-space distance margins as the primary open-set objective, 
\ourmethod\ introduces a dedicated $(K+1)$-way classification mechanism. 
This framework ensures that for any node $v_i \in \mathcal{V}_{test}$, the prediction is derived via:
\begin{equation}
    \hat{y}_i = \arg\max_{c \in \{1, \dots, K, K+1\}} \mathbf{P}(y_i = c \mid \mathbf{x}_i, \mathbf{A}),
\end{equation}
where $c = K+1$ represents the dynamic rejection slot for pseudo-unknown node variations synthetically extrapolated along heterophilic structural directions. 

\section{Methodology}
\label{sec:methodology}
\begin{figure*}
\centering
\includegraphics[width=.95\linewidth]{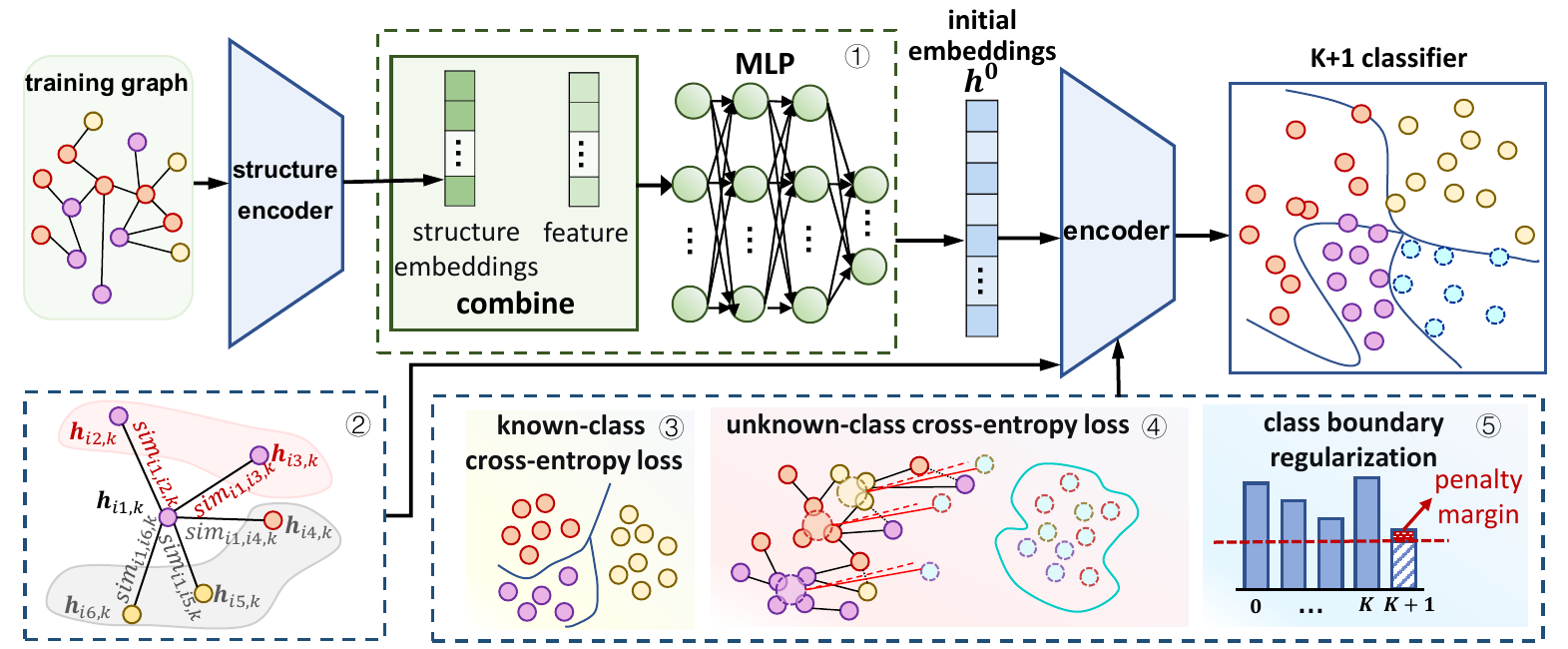}
\caption{The overall architecture of \ourmethod. Raw node features are enriched with structural encodings into initial representations $\mathbf{h}_0$. The structure encoder then extracts multi-hop structural patterns through selective neighborhood pathways to output enhanced intermediate embeddings. Finally, a unified $(K+1)$-way classifier categorizes nodes, jointly optimized via a multi-task learning paradigm comprising known-class loss $\mathcal{L}_{real}$, pseudo-extrapolated unknown-class loss $\mathcal{L}_{syn}$, and classification-driven regularization penalty $\mathcal{L}_{reg}$ without geometric margin contractions.} \label{figs:overview}
\end{figure*}

In this section, we elaborate on the architectural design of \ourmethod, a novel framework tailored for semi-supervised open-set node classification on heterophilic graphs. 
As shown in Fig.~\ref{figs:overview}, \ourmethod\ consists of four components: a structure-augmented feature initialization layer, a trustworthy neighborhood aggregation mechanism, a classification loss guided by a structural pseudo-extrapolation strategy, and a joint classification optimization framework integrated with logit margin regularization. 
By dynamically generating out-of-distribution (OOD) node boundaries without relying on brittle margin-based constraints, \ourmethod\ effectively prevents the overlapping of known and unseen class distributions under severe structural heterophily.

\subsection{Structure-Augmented Feature Initialization}
In heterophilic open-set graphs, known and unknown nodes are often structurally mixed, making it difficult to distinguish unknown classes after neighborhood aggregation. 
Therefore, the model should establish discriminative node representations before the message passing stage. 
An appropriate initial representation should capture both semantic attributes and topological structures, because in heterophilic graphs, topological connections carry important boundary patterns that can indicate whether a node lies in a mixed neighborhood containing open-set nodes.
To construct a complete initial node representation and avoid structural confusion at the beginning of the network, we enrich the raw feature space by explicitly adding structural encodings. This design ensures that the model recognizes structural identities at open-set boundaries before feature propagation.

For each node $v_i$, we pre-compute a 16-dimensional topology-only structural encoding to capture its multi-hop structural characteristics before message passing. Specifically, let $\mathbf{P}=\mathbf{A}\mathbf{D}^{-1}$ denote the degree-normalized propagation matrix, where $\mathbf{D}$ is the degree matrix. We define the structural encoding as
\[
\mathbf{s}_i=
\left[
(\mathbf{P})_{ii},
(\mathbf{P}^{2})_{ii},
\ldots,
(\mathbf{P}^{16})_{ii}
\right]
\in\mathbb{R}^{16}.
\]
The $k$-th component $(\mathbf{P}^{k})_{ii}$ characterizes the structural return pattern of node $v_i$ after $k$ propagation steps, allowing $\mathbf{s}_i$ to summarize local topology over multiple neighborhood ranges. This encoding depends only on graph structure, requires no node labels, and is shared across different backbone choices.

We then combine the structural encoding with the original node attributes to obtain the initial representation:
\begin{equation}
    \mathbf{h}^{(0)}_i
    =
    \operatorname{MLP}_{\mathrm{init}}
    \left(
    [\mathbf{x}_i \Vert \mathbf{s}_i]
    \right),
\end{equation}
where $\Vert$ denotes feature concatenation. This structure-augmented representation provides topology-aware initialization for the subsequent heterophily-aware message passing.

\subsection{Trustworthy Neighborhood Aggregation}
In heterophilic open-set graphs, not all neighbors provide useful information, as cross-class and open-set connections may introduce misleading messages during feature propagation. 
However, standard low-pass aggregation unconditionally averages all adjacent representations, which allows open-set node variants to indiscriminately poison the surrounding known-class representations across heterophilic links. 

To alleviate this problem, we design a trustworthy neighborhood aggregation module to selectively aggregate messages from reliable, contextually consistent neighbors, which shields the ego-node from noisy or conflicting semantic categories. 
By implementing a trustworthy neighborhood filtering mechanism, edge compatibility is evaluated to ensure that only highly reliable homophilic neighbors participate in the final neighborhood aggregation. Concretely, we utilize an edge discriminator to estimate the structural relationship between connected nodes. For each incoming edge $(v_j, v_i)$, the discriminator computes a homophily probability $p_{ij}$. Meanwhile, we compute the cosine similarity between the intermediate hidden states of node $v_i$ and node $v_j$. 
Let $\mathbf{h}_i^{(t-1)}$ denote the hidden embedding of node $v_i$ at the $(t-1)$-th iteration step. The model evaluates a homophily selection score $s_{ij}$ for each connection:
\begin{equation}
    s_{ij} = p_{ij} \cdot \frac{\text{ReLU}(\text{Sim}(\mathbf{h}_j^{(t-1)}, \mathbf{h}_i^{(t-1)}))}{\tau},
\end{equation}
where $\text{Sim}(\cdot)$ is the cosine similarity function, and $\tau$ represents the aggregation temperature. 
Rather than keeping all neighbors, a strict masking threshold is applied based on $s_{ij}$. An edge is preserved only if its homophily selection score is sufficiently high. This step filters out untrustworthy cross-class neighbors and extracts a clean, homophilic neighbor subset $\mathcal{N}_{trust}(v_i)$.

Finally, the model performs normalized aggregation exclusively over this trustworthy homophilic subset $\mathcal{N}_{trust}(v_i)$. The layer-wise representation update for node $v_i$ at the $t$-th step is formulated as follows:
\begin{equation}
\begin{aligned}
    \mathbf{h}_i^{(t)} = \text{LayerNorm} \bigg( 
     \text{MLP}_{fuse} \Big( \big[ \, \mathbf{h}_i^{(t-1)} \parallel {} \\
     \sum_{v_j \in \mathcal{N}_{trust}(v_i)} \alpha_{ij} \cdot \mathbf{h}_j^{(t-1)} \, \big] \Big) 
    + \mathbf{W}_{self} \cdot \mathbf{h}_i^{(0)} \bigg),
\end{aligned}
\end{equation}
where $\parallel$ denotes the concatenation operation, $\alpha_{ij}$ represents the normalized edge weight computed via Softmax over the trusted subset, and $\mathbf{W}_{self} \cdot \mathbf{h}_i^{(0)}$ provides a self-loop residual connection from the initialization layer. After $T$ iterations, the final output representation is denoted as $\mathbf{z}_i = \mathbf{h}_i^{(T)}$ for all $v_i \in \mathcal{V}$.
For different backbones, the backbone-specific propagation first produces a graph-aware representation, which is combined with $\mathbf{h}^{(0)}$ through a residual connection and subsequently refined by the same trustworthy aggregation module. Thus, HOPE does not alter the internal propagation rule of the underlying backbone.

Overall, this module enables \ourmethod to perform more robust and discriminative message passing by preserving beneficial homophilic signals while suppressing misleading heterophilic noise, thereby improving open-set recognition in complex graph structures.

\subsection{Pseudo-Unknown Proxy Generation and Classification Loss}
In open-set recognition, the absence of labeled unknown samples makes it difficult for the model to explicitly learn where known-class decision boundaries should stop. 
To bridge this gap, we synthesize representative out-of-distribution (OOD) boundary instances to explicitly populate the $(K+1)$-th classification slot, forcing the open-set model to learn a compact and closed decision boundary for known classes within a standard classification framework.
Under heterophilic environments, graph nodes generally exhibit a low edge homophily ratio $h$, meaning that their neighborhoods $\mathcal{N}(v_i)$ contain substantial semantic diversity across multiple categories.
Crucially, this structural characteristic applies to all entities in the graph, including both known and unknown classes.
During message-passing propagation, when a Graph Neural Network (GNN) aggregates multi-hop structural patterns, the features of an open-set node are simultaneously subjected to multi-directional traction exerted by its semantically diverse neighbors from distinct known categories.
This omnidirectional structural pulling prevents unknown nodes from forming isolated, well-segregated clusters in the latent space. 
Instead, their latent representations are inherently driven into the intersecting zones, peripheral margins, and ambiguous boundaries of the established known manifolds.

Consequently, the latent representations of nodes near these frontiers, regardless of whether they belong to known or unseen categories, would suffer from severe territorial overlap, which blurs the closed-set rejection boundaries.
To decouple this overlap without modifying or distorting the internal representation spaces of known classes, \ourmethod leverages a structural extrapolation paradigm optimized within mini-batches. 
Instead of forcing the learned clusters of seen categories to become overly compact, our method places synthetic boundary proxies in the ambiguous regions between different classes.
By assigning these boundary proxies to the unique classification slot $c_{unk} = K+1$, the standard cross-entropy loss forces the $(K+1)$-th logit to become dominant exactly within these overlapping regions.
As a result, the linear decision boundaries of the classifier are driven to adaptively wrap around and seal the known manifolds, effectively delegating the contaminated intersection zones to the unknown slot while leaving the internal latent structures of known classes uncompromised and largely preserved. 

Specifically, during each training epoch, we optimize HOPE on the subgraph induced by labeled training nodes. To provide stable class anchors, we maintain an EMA center $\boldsymbol{\mu}_c\in\mathbb{R}^d$ for each known class $c\in\mathcal{Y}_L$:
\begin{equation}
    \boldsymbol{\mu}_c
    \leftarrow
    \rho\boldsymbol{\mu}_c
    +(1-\rho)
    \frac{1}{|\mathcal{V}_{\mathrm{train}}^c|}
    \sum_{v_i\in\mathcal{V}_{\mathrm{train}}^c}
    \mathbf{z}_i ,
\end{equation}
where $\mathcal{V}_{\mathrm{train}}^c$ denotes the labeled training nodes belonging to class $c$, and $\rho$ is the EMA smoothing factor.

To synthesize pseudo-unknown proxies, we sample structurally ambiguous known-class nodes as anchors according to a softened structural score that combines neighborhood entropy and local cross-class connectivity. All statistics involved in anchor selection are computed exclusively on the labeled training-induced subgraph.

For an anchor node $v_i$ with label $y_i$, we define its known heterophilic neighborhood as
$\mathcal{N}_{\mathrm{het}}(v_i)=
\{v_j\in\mathcal{N}(v_i)\cap\mathcal{V}_{\mathrm{train}}
\mid y_j\neq y_i,\; y_j\in\mathcal{Y}_L\}$.
We then construct an outward extrapolation direction from the center of the anchor class toward its heterophilic neighbors:
$\mathbf{d}_i=
\frac{1}{|\mathcal{N}_{\mathrm{het}}(v_i)|}
\sum_{v_j\in\mathcal{N}_{\mathrm{het}}(v_i)}
(\mathbf{z}_j-\boldsymbol{\mu}_{y_i})$.
Accordingly, the pseudo-unknown representation is generated as
\begin{equation}
    \tilde{\mathbf{z}}
    =
    \boldsymbol{\mu}_{y_i}
    +
    \beta\mathbf{d}_i
    +
    \boldsymbol{\epsilon},
\end{equation}
where $\beta\sim\mathcal{U}(1,\beta_{\max})$ and
$\beta_{\max}=\max(1,\beta_{\mathrm{base}}(1+\eta))$.
Here, $\eta$ denotes the heterophily ratio of known-class edges in the current training-induced subgraph, which adaptively controls the extrapolation range, while
$\boldsymbol{\epsilon}\sim\mathcal{N}(\mathbf{0},\sigma^2\mathbf{I})$
introduces mild perturbations to improve boundary coverage.

To jointly preserve known-class discrimination and learn the additional rejection slot, we optimize real known nodes and synthesized pseudo-unknown proxies in a unified $(K+1)$-way classification space. For a labeled training node
$v_i\in\mathcal{V}_{\mathrm{train}}$ with $y_i\in\mathcal{Y}_L$, we apply the supervised cross-entropy loss only over the first $K$ known-class logits:
\begin{equation}
    \mathcal{L}_{\mathrm{real}}
    =
    -\frac{1}{|\mathcal{V}_{\mathrm{train}}|}
    \sum_{v_i\in\mathcal{V}_{\mathrm{train}}}
    \log
    \frac{\exp(\mathbf{o}_{i,y_i})}
    {\sum_{c=1}^{K}\exp(\mathbf{o}_{i,c})},
\end{equation}
where
$\mathbf{o}_i=\operatorname{Linear}(\mathbf{z}_i)\in\mathbb{R}^{K+1}$
is the output logit vector and $\mathbf{o}_{i,c}$ denotes its $c$-th component.

Concurrently, we optimize the generated proxies toward the $(K+1)$-th unknown slot using a weighted cross-entropy loss:
\begin{equation}
    \mathcal{L}_{\mathrm{syn}}
    =
    -\frac{1}{\sum_{q=1}^{|\mathcal{V}_{\mathrm{syn}}|} w_q}
    \sum_{q=1}^{|\mathcal{V}_{\mathrm{syn}}|}
    w_q
    \log
    \frac{\exp(\tilde{\mathbf{o}}_{q,K+1})}
    {\sum_{c=1}^{K+1}\exp(\tilde{\mathbf{o}}_{q,c})},
\end{equation}
where $\tilde{\mathbf{o}}_q=\operatorname{Linear}(\tilde{\mathbf{z}}_q)$, and $w_q$ denotes the structural sampling weight inherited from the corresponding anchor.

\subsection{Logit Margin Regularization}

To prevent the unknown slot from dominating the logit space of known-class nodes during training, we introduce a classification-driven margin penalty $\mathcal{L}_{reg}$. This regularization term explicitly encourages that for any known-class training node $v_i$ (with $y_i \in \mathcal{Y}_L$), the logit corresponding to the unknown rejection slot $c_{unk}=K+1$ remains strictly lower than the maximum logit among the known classes. 
Without such a constraint, the model might push the unknown logit excessively high in order to fit synthetic pseudo-unknown proxies, thereby eroding the discriminative power of known classes.
Formally, the logit margin regularization loss $\mathcal{L}_{reg}$ enforces a margin between the strongest known-class logit and the unknown-class logit for each known training node: 

\begin{equation}
\begin{aligned}
    \mathcal{L}_{reg} = \frac{1}{|\mathcal{V}_{train}|} \sum_{v_i \in \mathcal{V}_{train}} \Big[ & \max\bigl(0,\; \mathbf{o}_{i,K+1} - \\
    & \max_{c=1,\dots,K} \mathbf{o}_{i,c} + m \bigr) \Big],
\end{aligned}
\end{equation}
where $\mathcal{V}_{train}^{known}$ is the set of training nodes whose labels belong to the known classes $\mathcal{Y}_L$, $m>0$ is a fixed margin hyperparameter, and $\max(0,\cdot)$ denotes the ReLU function. 
This loss penalizes a known-class node whenever its maximum known-class logit fails to exceed the unknown-slot logit by at least the margin $m$, thereby encouraging a clear separation between known-class predictions and the rejection option. 

Without this regularization, the joint optimization of $\mathcal{L}_{real}$ and $\mathcal{L}_{syn}$ may inadvertently drive the unknown logit to become active even for known nodes, especially in heterophilic graphs where known and unknown neighborhoods are heavily intertwined. 
By imposing a soft margin constraint directly on the logits, rather than on geometric distances in the representation space. \ourmethod avoids brittle boundary tuning while preserving the full expressiveness of the $(K+1)$-way classifier.

Finally, the total objective function of \ourmethod\ is formulated as a multi-task learning paradigm:
\begin{equation}
    \mathcal{L}_{total} = \mathcal{L}_{real} + \gamma_1 \mathcal{L}_{syn} + \gamma_2 \mathcal{L}_{reg},
\end{equation}
where $\gamma_1$ and $\gamma_2$ are non-negative hyperparameters that scale the contributions of the pseudo-unknown classification risk and the known-class logit regularization penalty, respectively. 
Overall, this unified objective allows \ourmethod to jointly optimize known-class discrimination, unknown-boundary modeling, and known-class rejection regularization, leading to more reliable open-set recognition under heterophilic graph structures. 

\section{Experiments}
\label{sec:experiments}

\subsection{Experimental Setup}
\label{sec:exp_setup}

\begin{table}[t]
\caption{Detailed statistics of the evaluated heterophilous graph datasets.}
\label{tab:dataset_stats}
\centering 
\footnotesize 
\setlength{\tabcolsep}{6pt} 

\begin{tabular}{lcccc}
\toprule
\textbf{Dataset} & \multicolumn{4}{c}{\textbf{Graph Properties}} \\ 
\cmidrule{2-5} 
\textbf{Name} & \textbf{Nodes} & \textbf{Edges} & \textbf{Features} & \textbf{Classes} \\
\midrule
Chameleon      & 2,277   & 31,421    & 2,325 & 5  \\
Squirrel       & 5,201   & 198,493   & 2,089 & 5  \\
Wisconsin      & 251     & 466       & 1,703 & 5  \\
Amazon-Ratings & 24,492  & 93,050    & 300   & 5  \\
Roman-Empire   & 22,662  & 32,927    & 300   & 18 \\
Actor          & 7,600   & 26,752    & 932   & 5  \\
Arxiv-Year     & 169,343 & 1,166,243 & 128   & 5  \\
\bottomrule
\addlinespace 
\multicolumn{5}{l}{\parbox{0.95\columnwidth}{$^{\mathrm{a}}$The class space indicates $K$ observed known classes and $1$ unobserved unknown class.}} \\
\end{tabular}
\end{table}
\begin{table*}[htbp]
\centering
\caption{Experimental results (Acc and F1) across multi-datasets}
\label{tab:results}

\footnotesize 
\setlength{\tabcolsep}{3.5pt} 
\renewcommand{\arraystretch}{1.1} 
\resizebox{1\textwidth}{!}{
\begin{tabular}{ll|ccccccc|ccccccc}
\toprule
 & & \multicolumn{7}{c|}{\textbf{Accuracy (\%)}} & \multicolumn{7}{c}{\textbf{F1-Score (\%)}} \\
\textbf{Model} & \textbf{Method} & 
\makecell[c]{\textbf{Roman}\\\textbf{Empire}} & \makecell[c]{\textbf{Amazon}\\\textbf{Ratings}} & \makecell[c]{\textbf{Wis-}\\\textbf{consin}} & \textbf{Actor} & \makecell[c]{\textbf{Chame-}\\\textbf{leon}} & \textbf{Squirrel} & \makecell[c]{\textbf{Arxiv-}\\\textbf{Year}} & 
\makecell[c]{\textbf{Roman}\\\textbf{Empire}} & \makecell[c]{\textbf{Amazon}\\\textbf{Ratings}} & \makecell[c]{\textbf{Wis-}\\\textbf{consin}} & \textbf{Actor} & \makecell[c]{\textbf{Chame-}\\\textbf{leon}} & \textbf{Squirrel} & \makecell[c]{\textbf{Arxiv-}\\\textbf{Year}} \\
\midrule

\multirow{6}{*}{\textbf{GCN}}
 & GCN & 20.58 & 31.63 & 47.76 & 12.48 & 43.24 & 14.68 & 44.91 & 16.13 & 11.20 & 29.57 & 12.35 & 36.57 & 22.35 & 32.23 \\
 & ROG\_PL & 35.15 & 31.68 & 52.94 & 25.48 & 33.04 & 21.64 & 41.52 & 29.48 & 19.68 & 19.66 & 19.90 & 15.18 & 15.83 & 28.21 \\
 & G2Pxy & 31.60 & 30.85 & 41.88 & 20.43 & 31.92 & 18.55 & 42.62 & 37.01 & 21.98 & 33.72 & 13.23 & 18.26 & 20.01 & 30.22 \\
 & EGonc & 34.24 & 37.54 & 49.14 & 17.43 & 46.27 & 15.98 & 40.98 & 29.57 & 22.67 & 19.23 & 18.70 & 31.44 & 16.77 & 30.60 \\
 & CONC & 28.72 & 31.38 & 42.64 & 11.76 & 15.53 & 17.02 & 38.60 & 15.32 & 11.47 & 13.16 & 7.48 & 18.38 & 19.10 & 29.66 \\
 & \ourmethod & \textbf{53.95} & \textbf{39.92} & \textbf{59.32} & \textbf{43.85} & \textbf{55.16} & \textbf{39.00} & \textbf{45.41} & \textbf{54.78} & \textbf{55.28} & \textbf{43.09} & \textbf{28.87} & \textbf{38.76} & \textbf{21.29} & \textbf{35.85} \\
\midrule

\multirow{6}{*}{\textbf{GPR-GNN}}
 & GPR-GNN & 40.05 & 30.43 & 49.52 & 11.97 & 43.08 & 16.44 & 42.39 & 33.81 & 12.52 & 32.08 & 12.21 & 36.49 & 23.87 & 28.33 \\
 & ROG\_PL & 47.14 & 32.28 & 47.05 & 28.13 & 24.50 & 26.82 & 34.79 & 41.07 & 17.26 & 20.60 & 21.19 & 21.34 & 17.27 & 20.26 \\
 & G2Pxy & 42.53 & 30.74 & 40.68 & 22.09 & 27.44 & 19.72 & 32.17 & 27.34 & 21.76 & 34.29 & 16.33 & 18.26 & 20.23 & 19.77 \\
 & EGonc & 34.21 & 37.62 & 45.33 & 21.34 & 36.58 & 16.33 & 37.52 & 34.56 & 28.51 & 23.12 & 17.71 & 20.57 & 17.45 & 24.85 \\
 & CONC & 36.66 & 31.38 & 36.64 & 12.22 & 21.27 & 19.20 & 40.60 & 32.78 & 19.62 & 33.28 & 11.44 & 18.38 & 20.18 & 29.33 \\
 & \ourmethod & \textbf{50.04} & \textbf{38.81} & \textbf{59.32} & \textbf{42.22} & \textbf{56.99} & \textbf{46.21} & \textbf{44.58} & \textbf{51.04} & \textbf{33.80} & \textbf{41.44} & \textbf{27.69} & \textbf{39.93} & \textbf{29.53} & \textbf{32.01} \\
\midrule

\multirow{6}{*}{\textbf{GCNII}}
 & GCNII & 44.48 & 21.47 & 47.76 & \textbf{34.18} & 29.87 & 31.85 & 42.13 & 33.87 & 12.00 & 28.85 & 19.08 & 28.71 & 22.38 & 26.59 \\
 & ROG\_PL & 46.22 & 31.66 & 46.80 & 27.05 & 34.46 & 41.39 & 38.55 & 34.55 & 14.14 & 15.67 & 17.58 & 11.33 & 23.54 & 23.52 \\
 & G2Pxy & 45.54 & 29.19 & 47.06 & 28.64 & 42.56 & 40.76 & 35.23 & 36.27 & 25.22 & 24.20 & 11.21 & 22.60 & 23.62 & 22.17 \\
 & EGonc & 44.91 & 30.61 & 42.64 & 28.40 & 46.20 & 34.03 & 30.76 & 26.28 & 20.53 & 21.05 & 17.76 & 23.74 & 21.77 & 22.01 \\
 & CONC & 33.65 & 32.05 & 42.21 & 25.64 & 33.54 & 34.13 & 36.61 & 24.61 & 22.34 & 26.87 & 12.09 & 28.18 & 22.14 & 25.40 \\
 & \ourmethod & \textbf{51.51} & \textbf{38.67} & \textbf{55.93} & \underline{30.95} & \textbf{48.16} & \textbf{42.84} & \textbf{43.99} & \textbf{52.06} & \textbf{32.69} & \textbf{38.73} & \textbf{29.15} & \textbf{38.04} & \textbf{27.60} & \textbf{29.96} \\
\midrule

\multirow{6}{*}{\textbf{EG-GCN}}
 & EG-GCN & 52.66 & 21.89 & \textbf{62.50} & 40.76 & 52.77 & 41.03 & 45.52 & 48.64 & 22.01 & \textbf{50.24} & 21.87 & 38.79 & 21.63 & 33.63 \\
 & ROG\_PL & 49.62 & 31.38 & 49.15 & 18.02 & 44.07 & 38.89 & 37.90 & 33.55 & 17.14 & 22.83 & 9.97 & 14.34 & 22.33 & 29.33 \\
 & G2Pxy & 47.43 & 28.71 & 37.50 & 22.09 & 41.22 & 39.48 & 40.28 & 27.74 & 23.12 & 12.00 & 16.36 & 28.61 & 22.76 & 30.24 \\
 & EGonc & 34.43 & 28.17 & 42.76 & 21.34 & 38.36 & 34.13 & 34.55 & 24.84 & 20.42 & 25.32 & 17.73 & 29.65 & 21.77 & 26.86 \\
 & CONC & 31.98 & 27.92 & 40.67 & 12.22 & 50.54 & 26.82 & 41.36 & 33.44 & 16.33 & 11.56 & 17.62 & 26.27 & 24.75 & 30.86 \\
 & \ourmethod & \textbf{53.13} & \textbf{29.33} & \underline{53.70} & \textbf{43.94} & \textbf{53.07} & \textbf{43.22} & \textbf{47.62} & \textbf{51.78} & \textbf{25.71} & \underline{40.13} & \textbf{26.73} & \textbf{46.23} & \textbf{25.02} & \textbf{33.97} \\
\bottomrule
\end{tabular}}
\end{table*}

To comprehensively evaluate the performance of our proposed framework, we conduct benchmark evaluations across seven widely-used heterophilic graph datasets, namely Roman-Empire \cite{platonov2023critical}, Amazon-Ratings \cite{platonov2023critical}, Wisconsin \cite{pei2020geom}, Actor \cite{tang2009social}, Chameleon \cite{pei2020geom,rozemberczki2021multi}, Squirrel \cite{pei2020geom,rozemberczki2021multi}, and Arxiv-Year \cite{lim2021new}. These benchmarks span diverse topological scales, attribute dimensionalities, and feature densities, providing a robust and challenging testbed for open-set node classification under heterophilic environments. Following standard semi-supervised open-set evaluation protocols established in graph domains, we designate the specific category with the fewest instances as the unobserved unknown novel class, ensuring it remains completely inaccessible during the optimization phase, while the remaining classes constitute the observed known label space. The source code and experimental configurations are available at:
\url{https://github.com/Solkattkgo/HOPE}.

As a plug-and-play framework adaptable to standard backbones, we evaluate \ourmethod\ by comparing it against two representative groups of state-of-the-art graph baselines. The first group consists of graph neural network architectures including GCN \cite{kipf2016semi}, GPR-GNN \cite{chien2020adaptive}, GCNII \cite{chen2020simple}, and EG-GCN \cite{liu2025integrating}, which serve as foundational structural encoders covering both homophilous assumptions and heterophilic designs. The second group comprises open-set detection frameworks including ROG\_PL \cite{zhang2024rog_pl}, G2Pxy \cite{pxy_zhang2023g2pxy}, EGonc \cite{zhang2024egonc} and CONC \cite{zhang2024conc}, which represent advanced open-set node classification and boundary modeling methods. Crucially, for the specific evaluations conducted in the open-set and closed-set performance analysis, robustness evaluation, and computational efficiency analysis, GCNII, GPR-GNN, and EG-GCN utilize their own specialized architectures as structural encoders, whereas all other compared open-set detection methods consistently adopt GCN as their default underlying backbone network.

To evaluate open-set node classification, we use Overall Accuracy and Macro-F1-Score. Overall Accuracy measures global prediction performance, while Macro-F1 provides a balanced evaluation across imbalanced classes. Unless otherwise specified, we set $\gamma_1=0.5$, $\gamma_2=0.1$, and $m=0.3$. The random seed is fixed to 42 for all randomized operations and dataset splits.

\subsection{Performance Comparison}
\label{sec:main_results}

The comparison results of \ourmethod\ are illustrated in Table~\ref{tab:results}. From the table, we make the following key observations:

Our proposed \ourmethod\ consistently secures the optimal or competitive second-best performance across almost all evaluation slots under both metrics. When paired with standard backbones, \ourmethod\ yields substantial performance gains compared to the vanilla versions. For instance, on the Roman-Empire dataset under the GPR-GNN framework, our method elevates the accuracy from 40.05\% to the best 50.04\% and the macro-F1 score from 33.81\% to the best 51.04\%. This performance stability highlights that our approach generalizes well to various heterophilic structural patterns.

The strong performance of \ourmethod\ is rooted in its dedicated design for open-set node recognition under severe structural heterophily. Specifically, it enriches raw features with multi-hop structural patterns via structure-augmented feature initialization, selectively propagates semantically consistent messages using a trustworthy neighborhood aggregation module, and synthesizes realistic pseudo-unknown boundaries via heterophily-guided structural extrapolation. Together with a joint classification framework with logit margin regularization, \ourmethod\ successfully circumvents severe statistical trade-offs, yielding balanced and stable leads.

In contrast, alternative baselines encounter clear algorithmic bottlenecks. Traditional closed-set encoders including GCN, GCNII, GPR-GNN, and EG-GCN lack native open-set awareness, and their confidence calibration breaks down under heterophilic linking patterns when evaluating under post-hoc threshold deployment protocols. 
This explains why, within the GCNII‑backbone group, the vanilla GCNII achieves the highest Overall Accuracy on the Actor dataset but exhibits a depressed Macro‑F1 score. 
On the other hand, established open-set baselines like ROG\_PL, G2Pxy, EGonc, and CONC assume structural homophily. When applied to heterophilic graphs, their rigid boundaries aggressively classify valid normal nodes into the open-set rejection slot due to severe false-positive errors. This over-rejection behavior accounts for the severe statistical trade-offs observed in the results, such as when ROG\_PL is paired with GPR-GNN on the Amazon-Ratings dataset, or combined with GCNII on the Squirrel dataset, where it achieves a relatively high Overall Accuracy while its Macro-F1 score lags far behind due to the catastrophic collapse of known-class precision.

\subsection{Ablation Study}
\label{sec:ablation}
\begin{table}[t]
\centering
\caption{Ablation study of the proposed model on Roman-Empire, Wisconsin, and Squirrel datasets.}
\label{tab:ablation_results}
\begin{tabular}{lcccccc}
\toprule
 & \multicolumn{2}{c}{Roman-Empire} & \multicolumn{2}{c}{Wisconsin} & \multicolumn{2}{c}{Squirrel} \\ 
\cmidrule(r){2-3} \cmidrule(lr){4-5} \cmidrule(l){6-7}
 & acc & f1 & acc & f1 & acc & f1 \\ 
\midrule
w/o reg    & 18.63 & 17.68 & 16.95 & 5.80 & 55.50 & 14.28 \\
w/o init    & 48.39 & 40.41 & 59.32 & 45.12 & 37.87 & 20.28 \\
w/o trust  & 47.16 & 48.17 & 57.63 & 40.52 & 38.51 & 20.64 \\
Full Model & 53.95 & 54.78 & 59.32 & 55.28 & 39.00 & 21.29 \\ 
\bottomrule
\end{tabular}
\end{table}

To examine the contribution of key designs in \ourmethod, we conduct ablation studies on 3 datasets with a GCN backbone. We compare the full model against three variants: w/o init, which removes the structure-augmented feature initialization layer; w/o trust, which disables the trustworthy neighborhood aggregation mechanism; and w/o reg, which omits the logit margin regularization loss. We have the following observations from Table~\ref{tab:ablation_results}: \ding{182}~The full model achieves the best overall performance, demonstrating that all designed modules are essential and mutually reinforcing.
\ding{183}~Removing the structural initialization layer triggers a significant performance degradation. This decline occurs because raw node attributes lack geometric awareness, making the model blind to local topological positions. These findings confirm that relying solely on semantic features is insufficient in heterophilic environments.
\ding{184}~Disabling the edge-filtering mechanism leads to noticeable performance drops across all datasets. The root cause is that standard low-pass aggregation unconditionally averages all adjacent representations, allowing open-set node variants and heterophilic cross-class neighbors to indiscriminately poison the ego-node features. 
\ding{185}~Omitting the logit margin regularization loss causes a catastrophic failure mode where the model collapses to classifying almost all nodes as the open-set class. This dramatic collapse verifies that our classification-driven margin penalty is indispensable for anchoring known-class logits to sustain a stable open-set decision space.

\subsection{Open-set and Closed-set Performance}
\label{sec:openset_closedset}

\begin{figure}[t]
    \centering
    \begin{subfigure}{0.49\linewidth}
        \centering
        \includegraphics[width=\linewidth, clip]{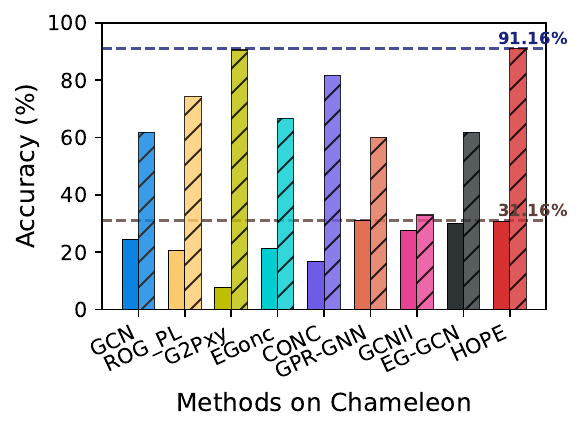}
        \caption{Chameleon}
        \label{fig:openset_cha}
    \end{subfigure}
    \begin{subfigure}{0.49\linewidth}
        \centering
        \includegraphics[width=\linewidth, clip]{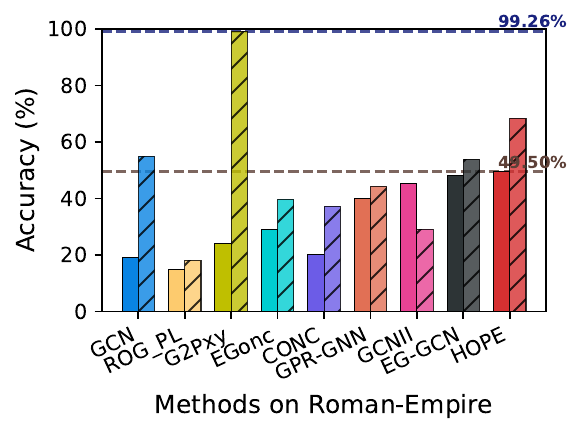}
        \caption{Roman-Empire}
        \label{fig:openset_rm}
    \end{subfigure}
    \caption{ACC of known/unknown classes comparison.}
    \label{fig:openset_comparison}
\end{figure}

To evaluate the fine-grained discriminative capability of \ourmethod\ in open-set scenarios, we analyze the classification performance on both known and unknown classes across the Chameleon and Roman-Empire datasets, as illustrated in Fig.~\ref{fig:openset_comparison}. 
From the results, we observe that \ourmethod\ exhibits an outstanding capability to simultaneously maintain high classification precision on observed known classes and achieve superior detection rates on unobserved unknown nodes. Although G2Pxy secures a higher individual accuracy for identifying unknown class nodes on the Roman-Empire dataset, it severely sacrifices the prediction accuracy of the normal known classes, leading to massive false-positive classification errors. In contrast, by utilizing structure-augmented initialization and trustworthy aggregation alongside balanced margin regularization, \ourmethod\ successfully manages the decision spaces and prevents the unknown slot from aggressively absorbing normal nodes, thereby establishing stable and comprehensive leads under severe structural heterophily.

\subsection{Robustness Analysis}
\label{sec:robustness}

\begin{figure}[t]
    \centering
    \begin{subfigure}{0.49\linewidth}
        \centering
        \includegraphics[width=\linewidth]{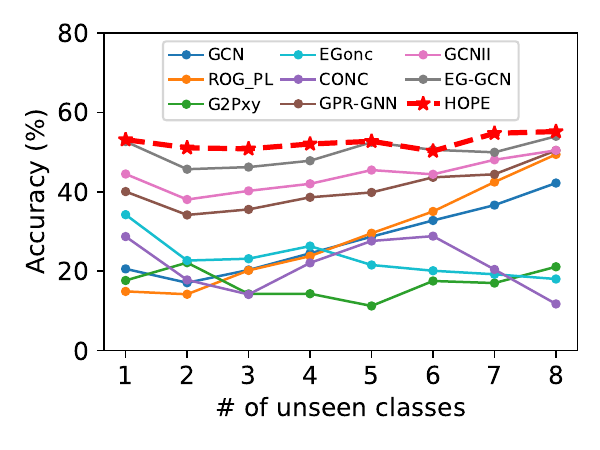}\vspace{-2mm}
        \caption{Accuracy comparison}
        \label{fig:robustacc}
    \end{subfigure}
    \begin{subfigure}{0.49\linewidth}
        \centering
        \includegraphics[width=\linewidth]{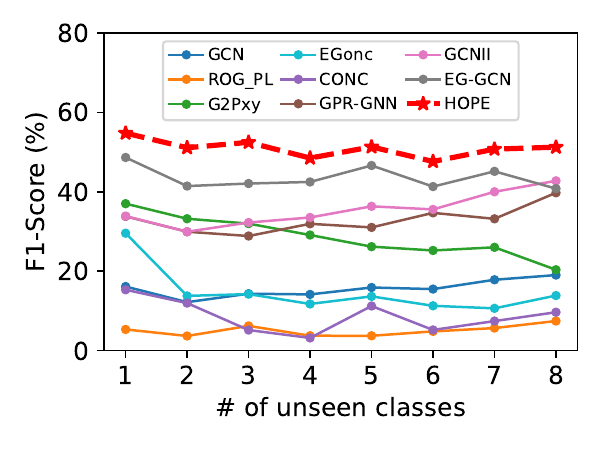}\vspace{-2mm}
        \caption{F1-score comparison}
        \label{fig:robustf1}
    \end{subfigure}
    \caption{ACC/F1 with the increasing \#open-set classes. }
    \label{fig:robust}
\end{figure}

To evaluate the operational stability and robustness of our proposed framework under volatile open-set environments, we show the performance variations on the Roman-Empire dataset by incrementally increasing the number of unseen classes from one to eight. 
As observed from the results in Fig.~\ref{fig:robust}, \ourmethod consistently maintains the optimal performance across all evaluation phases, exhibiting strong resistance against environmental volatility compared to alternative baselines. Even when the open-set class space expands during deployment, our framework establishes a steady performance superiority. This robust behavior is mainly attributed to our specialized pseudo-unknown proxy generation strategy, which models the invariant topological mixture mechanism within local mini-batches by adaptively shifting proxies outwards along heterophilic neighborhood displacement vectors. Consequently, our synthetic proxies continue to tend to populate representation regions associated with known-class boundaries to maintain stable decision hyperplanes under structural heterophily.
\subsection{Latent Space Topology Visualization}
\label{sec:visualization}
\begin{figure}
\centering
\includegraphics[width=.95\linewidth]{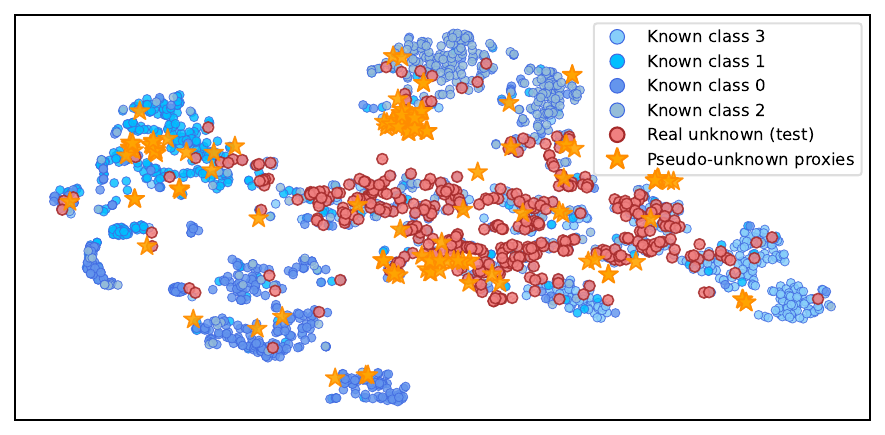}
\caption{The t-SNE visualization of node representations on the Chameleon dataset. The synthesized proxies tend to occupy ambiguous regions near known-class boundaries and show substantial overlap with regions containing real unknown nodes.}
\label{fig:tsne}
\end{figure}

To intuitively demonstrate the geometric soundness of our proposed structural pseudo-extrapolation strategy, we conduct a latent space topology visualization experiment using the t-SNE algorithm on the Chameleon dataset. The resulting visual distribution mapping is illustrated in Fig.~\ref{fig:tsne}. 
As shown in the t‑SNE visualization, the real unknown test nodes are distributed across specific regions adjacent to the known class boundaries. Importantly, rather than scattering randomly or encroaching upon the dense cores of known clusters, the synthetic boundary proxies tend to locate near the spatial positions occupied by the real unknown test nodes. This observation provides qualitative evidence that our proxies tend to appear near regions where real unknown nodes reside, offering visual support for the strategic rationality of our structural extrapolation paradigm under heterophily.

\subsection{Efficiency Analysis}
\label{sec:efficiency}
\begin{figure}[t]
    \centering
    \begin{subfigure}{0.48\linewidth}
        \centering
        \includegraphics[width=\linewidth]{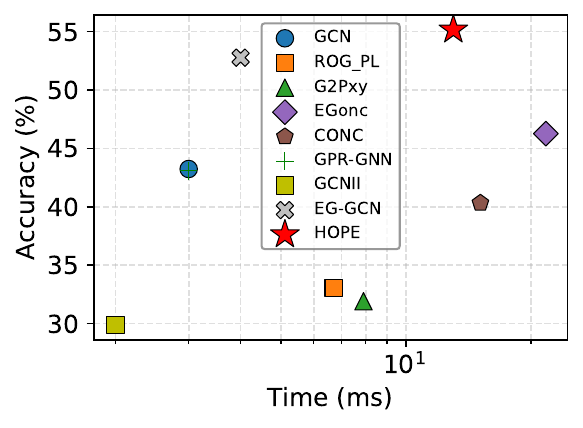}\vspace{-2mm}
        \caption{Accuracy vs. time usage}
        \label{fig:acc_time}
    \end{subfigure}
    \hfill
    \begin{subfigure}{0.48\linewidth}
        \centering
        \includegraphics[width=\linewidth]{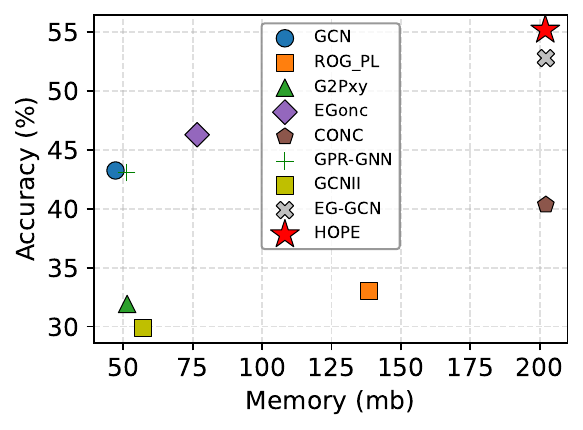}\vspace{-2mm}
        \caption{Accuracy vs. memory usage}
        \label{fig:acc_memory}
    \end{subfigure}
    \vspace{-2mm}
    \caption{Computational costs comparison.}
    \label{fig:efficiency}
\end{figure}

Complexity Analysis.
Let $N$, $M$, and $d$ denote the numbers of nodes, edges, and hidden dimensions, respectively. Excluding the structural encoding preprocessing, trustworthy edge scoring and aggregation require $O(Md)$ operations, while the edge MLP costs $O(Md^2)$ when its hidden width scales with $d$. Node transformations require $O(Nd^2)$, and proxy construction requires $O(Md+Sd)$ for $S$ synthesized proxies. Therefore, for fixed $d$ and $S$, the per-epoch complexity scales linearly with $N+M$, with memory complexity $O(Nd+Md)$.
Empirical Evaluation. We illustrate the execution trade-offs on the Chameleon dataset in Fig.~\ref{fig:efficiency}, where Fig.~\ref{fig:acc_time} and Fig.~\ref{fig:acc_memory} showcase accuracy versus time and memory usage, respectively. The plots indicate that alternative closed-set, open-set, and heterophilic baselines either suffer from poor performance or incur heavy computational overhead. In contrast, \ourmethod\ secures a substantial performance margin over these competitors while maintaining a small computational/memory footprint, achieving high accuracy with lower time and memory expenditures than heavy paradigms like EGonc.

\subsection{Parameter Sensitivity Analysis}
\label{sec:sensitivity}
\begin{figure}[t]
    \centering
    \begin{subfigure}{0.49\linewidth}
        \centering
        \includegraphics[width=\linewidth]{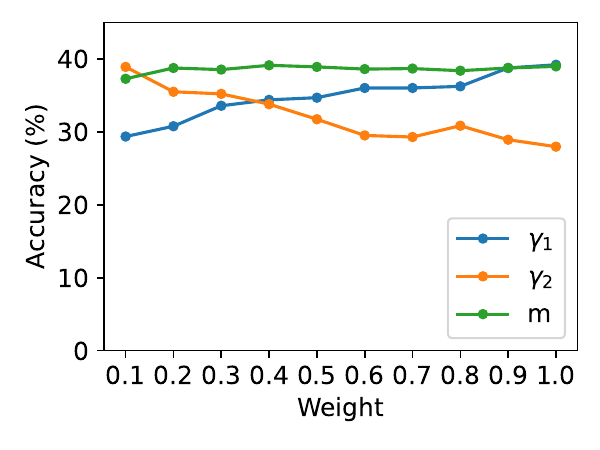}
        \caption{Sensitivity of Accuracy}
        \label{fig:sen_acc}
    \end{subfigure}
    \begin{subfigure}{0.49\linewidth}
        \centering
        \includegraphics[width=\linewidth]{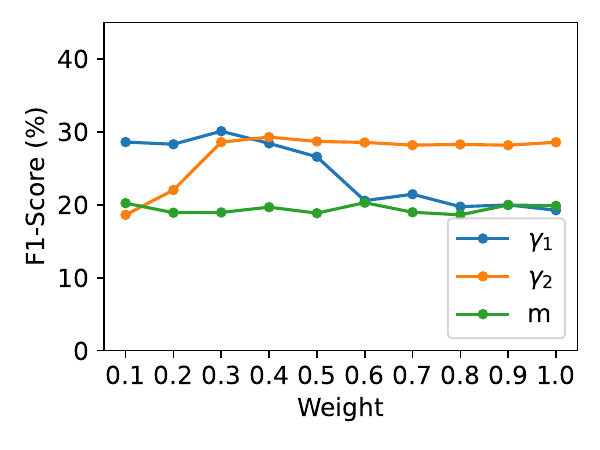}
        \caption{Sensitivity of F1 score}
        \label{fig:sen_f1}
    \end{subfigure}
    \caption{Parameter sensitivity analysis of \ourmethod with respect to $\gamma1, \gamma2$, and $m$ . The model shows consistent performance across a broad range of parameter settings.}
    \label{fig:sensitivity}
\end{figure}

To investigate the operational stability of \ourmethod\ under varied optimization balances, we perform a parameter sensitivity analysis on the Squirrel dataset. We evaluate the performance fluctuations in terms of Accuracy and Macro-F1 Score by varying the critical loss scaling factors $\gamma_1$, $\gamma_2$, and the soft margin $m$ independently within the range from 0.1 to 1.0. As illustrated in Fig.~\ref{fig:sensitivity}, different hyperparameter configurations present smooth and predictable performance variations. Specifically, $\gamma_1$ and $\gamma_2$ exhibit complementary tendencies due to the balance between synthetic open-set optimization and topological soft margin anchoring, while both metrics remain relatively stable and flat across the entire variation spectrum of the soft margin parameter $m$. Overall, HOPE remains stable across hyperparameter settings, demonstrating low sensitivity under heterophilic distributions. 

\section{Conclusion}
\label{sec:conclusion}

In this paper, we propose HOPE, a novel framework designed for open-set node classification on heterophilic graphs. To handle severe structural heterophily, our approach enriches raw features with multi-hop structural patterns and filters out noisy cross-class connections via a trustworthy aggregation mechanism. 
Furthermore, we introduce a heterophily-guided pseudo-extrapolation strategy to synthesize realistic pseudo-unknown boundaries at the intersections of known classes. 
This process is jointly optimized with a known-class logit regularization loss to maintain a balanced decision space. Extensive experiments across multiple benchmarks demonstrate that HOPE achieves the best or competitive performance across the evaluated settings against state-of-the-art baselines, verifying its effectiveness, efficiency, and robustness in entangled label environments.

\bibliographystyle{IEEEtran}
\bibliography{IEEEtran}

\end{document}